\def\year{2020}\relax
\documentclass[letterpaper]{article} % DO NOT CHANGE THIS
\usepackage{aaai20}  % DO NOT CHANGE THIS
\usepackage{times}  % DO NOT CHANGE THIS
\usepackage{helvet} % DO NOT CHANGE THIS
\usepackage{courier}  % DO NOT CHANGE THIS
\usepackage[hyphens]{url}  % DO NOT CHANGE THIS
\usepackage{graphicx} % DO NOT CHANGE THIS
\usepackage{graphicx}  % DO NOT CHANGE THIS
\usepackage{multirow}
\usepackage{amsmath}

\usepackage{algorithmic}

\usepackage{array}
\usepackage{amssymb}
\usepackage{ragged2e}
\usepackage{bm}
\title{Modality to Modality Translation: An Adversarial Representation Learning and Graph Fusion Network for Multimodal Fusion}
\author{ Sijie Mai, Haifeng Hu\thanks{Corresponding author}, Songlong Xing \\
\Large School of Electronics and Information Technology, Sun Yat-sen University\\ % All authors must be in the same font size and format. Use \Large and \textbf to achieve this result when breaking a line
\{maisj,xingslong\}@mail2.sysu.edu.cn, huhaif@mail.sysu.edu.cn\\ %If you have multiple authors and multiple affiliations
}
 
\begin{document}

\maketitle

\begin{abstract}
Learning joint embedding space for various modalities is of vital importance for multimodal fusion. Mainstream modality fusion approaches fail to achieve this goal, leaving a modality gap which heavily affects cross-modal fusion. In this paper, we propose a novel adversarial encoder-decoder-classifier framework to learn a modality-invariant embedding space. Since the distributions of various modalities vary in nature, to reduce the modality gap, we translate the distributions of source modalities into that of target modality via their respective encoders using adversarial training. Furthermore, we exert additional constraints on embedding space by introducing reconstruction loss and classification loss. Then we fuse the encoded representations using hierarchical graph neural network which explicitly explores unimodal, bimodal and trimodal interactions in multi-stage. Our method achieves state-of-the-art performance on multiple datasets. Visualization of the learned embeddings suggests that the joint embedding space learned by our method is discriminative.
\end{abstract}

\section{Introduction}\label{sec:Introduction}
Multimodal machine learning, where multiple data sources are available, always fares better compared with the situation where only single data source is utilized  \cite{8269806}. For example, sentiment analysis of text has been widely explored for a long time \cite{Liu2012}. But recent research shows that information from text is not sufficient for mining opinions of human \cite{Poria2017A}, especially under the situation where sarcasm or ambiguity is included. Nevertheless, if machine is able to understand a speaker's facial expression and how spoken language is uttered, it would be much easier to figure out the real sentiment. In this paper, we focus on analyzing human multimodal language from videos, where acoustic, language and visual modalities are included \cite{RMFN,MOSEI}. We can easily extend our work to other machine learning tasks as long as multiple modalities are presented, such as audio-visual speech recognition \cite{Su2018Multimodal} and cross-modal retrieval \cite{Xing2018Deep}.

How to fuse representations of various modalities has always been an open research. A key issue in multimodal fusion lies in the heterogeneous data distributions from different modalities \cite{8269806}, leading to the difficulty in mining complementary information across modalities which is critical for a comprehensive interpretation of multimodal information. The majority of previous works do not devote efforts to learn joint embedding space for various modalities to match multimodal distributions. Instead, they apply a subnetwork to each modality and then conduct fusion immediately \cite{Zadeh2017Tensor,Poria2017Context,Liu2018Efficient}. However, due to the heterogeneity across divergent modalities \cite{Lin2015Semantics}, the transformed representations learnt by these approaches can still follow unknown yet complex distributions.

Generative Adversarial Networks (GANs) \cite{GAN} have achieved significant progress, which can explicitly map one distribution to another prior distribution using adversarial training \cite{AAE}. Based on its unique characteristics, adversarial training is suitable for modality distribution translation. Motivated by it, we propose an adversarial encoder framework to match transformed distributions of all modalities and learn a modality-invariant embedding space. Specifically, source modalities' encoders, transforming the unimodal raw features into embedding space, try to fool the discriminator into distinguishing their encoded representations as those from target modality, while discriminator aims to classify the encoded representations from target modality as true but others as false. We also define one decoder for each modality that seeks to reconstruct the raw features to prevent unimodal information from being lost. Moreover, a classifier is built to classify the encoded representations into true category, which ensures that the embedding space is discriminative for learning task.

Furthermore, many prior methods fail to conduct fusion in a hierarchical way and are unable to explicitly model the interaction between each subset of multiple modalities \cite{Poria2017Context,Liu2018Efficient,Zadeh2018Multi}. In contrast, we interpret multimodal fusion as a hierarchical interaction learning procedure where firstly bimodal interactions are generated based on unimodal dynamics, and then trimodal dynamics are generated based on bimodal and unimodal dynamics. Drawing inspiration from recent success of graph-style neural network \cite{graph,MOSEI}, we propose a hierarchical fusion network, i.e., graph fusion network, which is highly interpretable in terms of the fusion procedure. Our graph fusion network consists of three layers containing unimodal, bimodal and trimodal dynamics respectively. The vertices in lower layer deliver their information to the higher layers where the information is fused to form multimodal information for the higher layers. By this means, we can explore cross-modal interactions hierarchically while still maintain the original interactions.

In brief, the main contributions are listed below:
\begin{itemize}
  \item We propose Adversarial Representation Graph Fusion framework (ARGF) for multimodal fusion. We address the importance of matching distributions before fusion and introduce adversarial training to learn a discriminative joint embedding space for various modalities, which can significantly narrow modality gap by matching multimodal distributions. Moreover, we define decoders and  classifier to retain unimodal information and enhance discrimination of the embedding space respectively.
  \item We propose a hierarchical graph fusion network which can explicitly model unimodal, bimodal and trimodal dynamics successively and is highly interpretable due to explicit meanings of vertices. It is flexible in fusion structure due to the changeable weights of edges and the ability to assign different importance to various interactions.
  \item ARGF achieves state-of-the-art performance on three multimodal learning datasets. We also provide visualizations for embedding space and graph fusion to prove their characteristics and contributions.
\end{itemize}
\section{Related Work}\label{sec:Related}
Earlier works on multimodal fusion focus on early fusion and late fusion. Early fusion approaches extract features of various modalities and fuse features at input level by simple concatenation \cite{Wollmer2013YouTube,Poria2017Convolutional}, but they cannot fully explore intra-modality dynamics as stated in \cite{Zadeh2017Tensor}.  In contrast, late fusion first makes decision according to each modality and then fuses the decisions  by weighted averaging \cite{Nojavanasghari2016Deep,Personality}, but it fails to model cross-modal interactions for features cannot interact with each other.

Recently, local fusion has become the mainstream strategy which can model time-dependent cross-modal interactions \cite{Gu2018Multimodal,Zadeh2018Multi,RAVEN}. For instance, \cite{Zadeh2018Memory} propose Memory Fusion Network that fuses memories of LSTMs over time, which is extended by \cite{MOSEI} using a Dynamic Fusion Graph (DFG) to fuse features. Our graph fusion network is different from DFG mainly in that: 1) we use inner product as part of edges' weight to estimate similarity between interactions; 2) in addition to fusing bimodal and unimodal dynamics, we also fuse each two bimodal dynamics to obtain more complete trimodal representations; 3) we determine the importance of vertices respectively to obtain each layer's output firstly rather than directly adding weighted information from all vertices. Tensor fusion has also become a new trend. Tensor Fusion Network \cite{Zadeh2017Tensor} adopts an outer product from multimodal embeddings to conduct fusion, followed by  \citeauthor{Liu2018Efficient} that tries to improve efficiency and reduces redundant information. More recently, \citeauthor{HFFN} propose a `Divide, Conquer and Combine' strategy to conduct local tensor and global fusion, which is extended in \cite{LMFN} using a more efficient feature segmentation method and an elaborately-designed BM-LSTM. In addition, some approaches use factorization methods to learn multimodal embedding \cite{MMB,MFM}. But methods above do not explicitly explore joint embedding space before fusion. Consequently, modality gap still seriously affects the effect of fusion.

For modality translation methods, Multimodal Transformer (MulT) \cite{MULT} transforms source modality to target modality using directional pairwise cross-modal attention. Moreover, Multimodal Cyclic Translation Network (MCTN) \cite{MCTN} learn joint representations via encoder-decoder framework by translating encoder's input (source modality) into decoder's output (target modality). MCTN uses encoder's output as joint representations of source and target modalities without exploring information from target modality, aiming to solve modality loss problem. In contrast, we address the problem of distribution translation and learn joint embedding space by translating the distributions of source modalities to target modality via adversarial training. Graph fusion is then conducted using encoded representations of both target and source modalities.

GANs \cite{GAN} have been successfully adopted in lots of learning tasks \cite{Xing2018Deep,Lin2015Semantics}. GANs have attracted significant interest in matching different distributions. Specifically, Adversarial Autoencoder \cite{AAE} aims to match aggregated posterior of hidden vector with a prior distribution. \cite{CM-GAN} seek to learn common representations for correlating heterogeneous data of various modalities, while we learn a modality-invariant embedding space for multimodal fusion by distribution translation. Besides, we add more constraints to transformed representations as well as separate the embedding space learning and cross-modal interaction learning procedures.

\section{Model Architecture}
\begin{figure*}
\centering
\includegraphics[width=1.9\columnwidth]{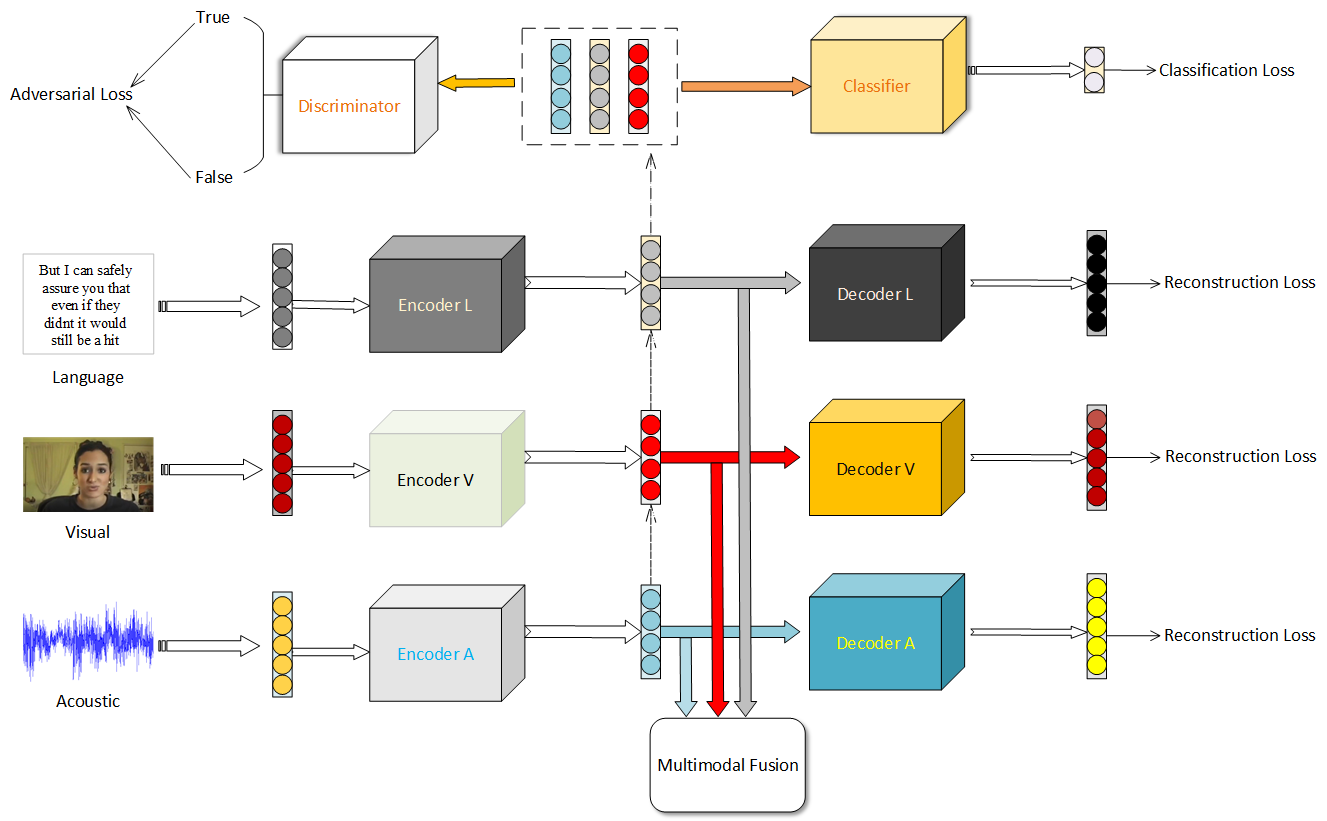}
\caption{\label{7}Schematic Diagram of ARGF. Input is a video of a speaker consisting of language, acoustic and visual modalities.
}
\end{figure*}

An overview of ARGF is given in Fig.~\ref{7}. ARGF is comprised of two stages: a joint embedding space learning stage and a graph fusion stage. In the first stage, we learn an embedding space for all modalities via adversarial framework. In the second stage, we utilize the representations output by encoders to conduct graph fusion (see Fig.~\ref{8}).

\subsection{Joint Embedding Space Learning}
In this section we provide an effective approach for translating distributions of source modalities to that of the target modality so as to learn a modality-invariant embedding space. Matching distributions in embedding space is crucial for modality fusion because the divergent statistical properties among modalities might seriously prevent cross-modal dynamics from being effectively explored. Drawing inspiration from GANs \cite{GAN}, we regularize the network by adversarial training. We also exert additional constraints like reconstruction loss and classification loss to optimize the learnt embedding space.

We start with notation: assume we have three modalities as input: acoustic $\bm{x}_a \in \mathbb{R}^{d_a}$, language $\ \bm{x}_l \in \mathbb{R}^{d_l}$ and visual $\bm{x}_v \in \mathbb{R}^{d_v}$ with $d_a$ being the dimensionality of $\bm{x}_a$ and so on. Assume that $\bm{x}_l$ is the target modality and other modalities are known as source modalities, and $p(\bm{x}_l)$ represents prior data distribution of language modality. Similar to \cite{AAE}, we define the transformed distributions of these three modalities as:
\begin{equation}
\begin{split}
 &p_{\theta_l}(\bm{\bm{x}_l^e})= \int_{\bm{x}_l}q(\bm{\bm{x}_l^e}|\bm{x}_l,\theta_l)p(\bm{x}_l)d\bm{x}_l\\
 &p_{\theta_a}(\bm{\bm{x}_a^e})= \int_{\bm{x}_a}q(\bm{\bm{x}_a^e}|\bm{x}_a,\theta_a)p(\bm{x}_a)d\bm{x}_a\\
&p_{\theta_v}(\bm{\bm{x}_v^e})= \int_{\bm{x}_v}q(\bm{\bm{x}_v^e}|\bm{x}_v,\theta_v)p(\bm{x}_v)d\bm{x}_v
 \end{split}
\end{equation}
where $q(\bm{\bm{x}_l^e}|\bm{x}_l,\theta_l)$ is known as the encoder function of language modality with $\theta_l$ being the parameters, and $p_{\theta_l}(\bm{\bm{x}_l^e})$ is the transformed language distribution in learnt embedding space restricted by $\theta_l$ (given specific input $\bm{x}_l,\ \bm{x}_a$ and $\bm{x}_v$, the encoded representations are $\bm{x}^e_l,\ \bm{x}^e_a$ and $\bm{x}^e_v$ respectively). Language encoder is a deep neural network that is denoted as $\bm{E}_l:$ $ \mathbb{R}^{d_l}\rightarrow \mathbb{R}^{k}$ for simplicity, where $k$ is the dimensionality of encoded representations, so are other encoders (all encoded representations share the same dimensionality $k$).

We hope that through optimizing $\theta_l,\ \theta_v$ and $\theta_a$, we can explicitly map transformed distributions $p_{\theta_a}(\bm{\bm{x}_a^e})$ and $p_{\theta_v}(\bm{\bm{x}_v^e})$ to $p_{\theta_l}(\bm{\bm{x}_l^e})$. However, the distributions of different modalities are very complex and they vary in nature which are extremely difficult to be matched by simple encoder networks. Therefore, we utilize adversarial training to add constraints to transformed distributions. Specifically, a discriminator $D$ is defined which aims to classify $p_{\theta_l}(\bm{\bm{x}_l^e})$ as true but $p_{\theta_a}(\bm{\bm{x}_a^e})$ and $p_{\theta_v}(\bm{\bm{x}_v^e})$ as false, while the generators (which are encoders $\bm{E}_a$ and $\bm{E}_v$) seek to fool discriminator $D$ into classifying $p_{\theta_v}(\bm{\bm{x}_v^e})$ and $p_{\theta_a}(\bm{\bm{x}_a^e})$ as true. The generators and discriminator compete with each other as a min-max game to learn joint embedding space. The loss function here can be divided into two parts: fake adversarial loss $\pounds_{fal}$ and true adversarial loss $\pounds_{tal}$, as shown below:
\begin{equation}
  \pounds_{al}=\pounds_{fal}(\bm{\bm{x}_a^e},\bm{\bm{x}_v^e}) +\pounds_{tal}(\bm{\bm{x}_l^e},\bm{\bm{x}_a^e},\bm{\bm{x}_v^e})
\end{equation}
$\bm{E}_a$ and $\bm{E}_v$ try to fool $D$ which results in $\pounds_{fal}$ while $D$ aims to determine the distribution of target modality as true but others as false, leading to $\pounds_{tal}$.  In practice, we define $\pounds_{fal}$ and $\pounds_{tal}$ as:
\begin{equation}
  \pounds_{fal}= -w [log(D(\bm{x}_a^e))+log(D(\bm{x}_v^e))]
\end{equation}
\begin{equation}
\pounds_{tal}= -w [log(1-D(\bm{x}_a^e))+log(1-D(\bm{x}_v^e))+ log(D(\bm{x}_l^e)) ]
\end{equation}
where $D(\bm{x}_a^e)$ denotes the predicted distribution value of $\bm{x}_a^e$  ranging from 0 to 1, and $w$ is the learnt weight for adversarial loss. If discriminator can not tell the target modality from all modalities (in the case where $D(\bm{x}_a^e)\!\approx\! D(\bm{x}_v^e)\!\approx\! D(\bm{x}_l^e)$), then the distributions of various modalities are successfully mapped into a modality-invariant embedding space. Adversarial training strategy puts restrictions on the statistical properties of the encoded representations. By adversarial training,  modality gap can be narrowed effectively so that representations from various modalities can be directly fused.

Transforming distributions might lead to the loss of unimodal information needed for mining complementary information between modalities. Therefore, to retain modality-specific information in learnt embedding space, we define decoders as:
\begin{equation}
\begin{split}
  &p_{\theta_{dl}}(\tilde{\bm{x}}_l)= \int_{\bm{x}^e_l} q(\tilde{\bm{x}}_l|\bm{x}^e_l,\theta_{dl})p_{\theta_l}(\bm{x}^e_l)d\bm{x}^e_l\\
  &p_{\theta_{da}}(\tilde{\bm{x}}_a)= \int_{\bm{x}^e_a} q(\tilde{\bm{x}}_a|\bm{x}^e_a,\theta_{da})p_{\theta_a}(\bm{x}^e_a)d\bm{x}^e_a\\
  &p_{\theta_{dv}}(\tilde{\bm{x}}_v)= \int_{\bm{x}^e_v} q(\tilde{\bm{x}}_v|\bm{x}^e_v,\theta_{dv})p_{\theta_v}(\bm{x}^e_v)d\bm{x}^e_v
\end{split}
\end{equation}
where $q(\tilde{\bm{x}}_l|\bm{x}^e_l,\theta_{dl})$ is the language decoder with $\theta_{dl}$ being the parameters, and $p_{\theta_{dl}}(\tilde{\bm{x}}_l)$ is the distributions of reconstructed representations for language modality. Given specific input $\bm{x}_f=\{\bm{x}_l,\ \bm{x}_a,\ \bm{x}_v\}$ to the encoders, we hope that the regenerated outputs of decoders $\tilde{\bm{x}}_f=\{\tilde{\bm{x}}_l,\tilde{\bm{x}}_a,\tilde{\bm{x}}_v\}$ approximate $\bm{x}_f$ to minimize information loss. To do so, we define a reconstruction loss as:
\begin{equation}
  \pounds_{rl}(\tilde{\bm{x}}_f,\bm{x}_f)\!\!=\!\!\sum_m\! \|\tilde{\bm{x}}_m-\bm{x}_m\|_2, m\! \in \! \{v,a,l\}
\end{equation}
By minimizing $\pounds_{rl}$, encoded representations can retain the unimodal information for further fusion.

Furthermore, to render the learnt embedding space more discriminative with respect to learning task, we also define a classifier which takes as input the encoded representations of each modality. The classifier is defined as:
\begin{equation}
\centering
   \tilde{\bm{y}}_m=\emph{C}(\bm{x}_{m}^e;\Theta_{c}),\ m\! \in \! \{v,a,l\}
\end{equation}
where $\emph{C}\in \mathbb{R}^{3k}\rightarrow \mathbb{R}^{N}$ denotes the classifier with $N$ being the number of categories to be classified, and $\tilde{\bm{y}}_m$ is the predicted label based on encoded representations of modality $m$. To minimize predicted error, we define a classification loss as:
\begin{equation}
  \pounds_{cl}(\tilde{\bm{y}},\bm{y})=\sum_m \|\tilde{\bm{y}}_m-\bm{y}\|_2
\end{equation}
where $\bm{y}$ is the true one-hot label. Classification loss enables the encoded representations to carry the needed label information and thus the embedding space is discriminative for learning tasks.

In conclusion, the total loss of this section is:
\begin{equation}
  \pounds = [\lambda\pounds_{fal}+ (1-\lambda)\cdot \pounds_{rl}] +0.5\pounds_{tal} + \pounds_{cl}
\end{equation}
where $\lambda$ is a hyper-parameter that determines the importance of loss functions whose value is determined by grid search. During gradient update, firstly, we apply $\pounds_{fal}$ and $\pounds_{rl}$ to update encoders and decoders; secondly, we use $\pounds_{tal}$ to update the discriminator to improve its discriminative power with respect to fake/true distributions; lastly, $\pounds_{cl}$ is utilized to update encoders and classifier to improve discrimination ability of joint embedding space.

\subsection{Graph Fusion Network}
\begin{figure}
\centering
\includegraphics[width=.95\columnwidth]{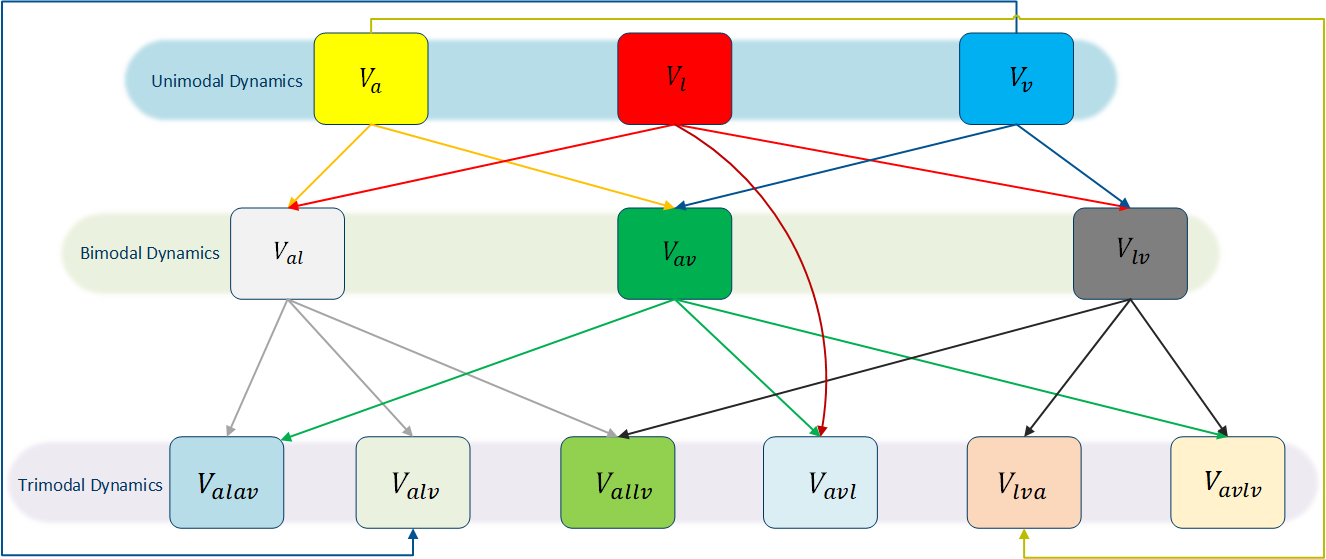}
\caption{\label{8}Schematic Diagram of GFN.
}
\end{figure}

We explore cross-modality interactions by fusing encoded representations of all modalities in this section. Assuming that multimodal fusion is in multi-stage \cite{RMFN} and considering the need to preserve all $n$-modal interactions, we introduce graph fusion network (GFN), a hierarchical neural network, to model unimodal, bimodal and trimodal interactions successively. As shown in Fig.~\ref{8}, the network consists of  unimodal, bimodal and trimodal dynamic learning layers. GFN regards each interaction as a vertex and the similarities between interacted vertices as well as the interacted vertices' importance as weights of edges, which is highly interpretable and flexible in terms of fusion structure.

The first layer is the unimodal dynamic learning layer consisting of unimodal vertices of three modalities whose information vectors are denoted as $\bm{V}_l,\ \bm{V}_a$ and $\bm{V}_v$, corresponding to the encoded representations $\bm{x}_l^e,\ \bm{x}_a^e$ and $\bm{x}_v^e$ respectively. In first layer, we apply a Modality Attention Network $MAN\!\in\! \mathbb{R}^{k}\! \rightarrow\! \mathbb{R}^{1}$ to process each vertex and determine the importance of these unimodal interactions since not all modalities contribute equally. Then we obtain final unimodal dynamics by calculating the weighted average of information from all unimodal vertices, as shown below:
\begin{equation}
\centering
   \begin{split}
   &\alpha_m = \emph{MAN}(\bm{V}_{m};\Theta_{MAN}) \\
   &\bm{U}=\frac{1}{3}\sum_m \alpha_m \cdot \bm{V}_{m},\ \ m \in \{ a,v,l\}
   \end{split}
\end{equation}
where $\bm{U}$ is the final unimodal vector and $\alpha_m$ is the weight of modality $m$. In practice, $MAN$ consists of a $sigmoid$-activated dense layer parameterized by $\Theta_{MAN}$.

In the second layer, i.e., bimodal dynamic learning layer, each two unimodal vertices are fused using a multi-layer neural fusion network: $MLP$ $\in \mathbb{R}^{2k} \rightarrow \mathbb{R}^{k}$ to obtain each bimodal vertex:
\begin{equation}
\centering
   \begin{split}
   &\bm{V}_{m_1m_2}=\emph{MLP}(\bm{V}_{m_1}\oplus \bm{V}_{m_2};\Theta_{MLP})\\
   &m_1,m_2\in \{a,v,l\},\ m_1 \ne m_2
   \end{split}
\end{equation}
where $\oplus$ denotes vector concatenation, $\bm{V}_{m_1m_2}$ is the information vector of $m_1m_2$ vertex and $\Theta_{MLP}$ is the parameter matrix of $MLP$. In practice, $MLP$ is composed of two dense layers activated by $Leaky\ ReLU$ and $tanh$ respectively. As for the weights of those edges linking the first and second layer, we firstly estimate the similarity of each two uniformed unimodal information vectors of first layer using inner product. We argue that the more similar two information vectors are, the less important their bimodal interaction would be. This is premised on the assumption that provided two information vectors are close to each other, then little complementary information lies between them and their information has been well explored in the first layer. The similarity of two information vectors is defined as:
\begin{equation}
\centering
   S_{m_1m_2}=\widetilde{\bm{V}}_{m_1}^T \widetilde{\bm{V}}_{m_2}
\end{equation}
where $\widetilde{\bm{V}}_{m_1}$ represents the softmax-normalized vector of $\bm{V}_{m_1}$ (the normalization ensures that the computed similarity is between 0 to 1), $T$ means vector transpose operation and $S_{m_1m_2}$ is a scalar that denotes the similarity of $m_1$ and $m_2$ vertices. The weight of edge that links $m_1$ vertex in first layer and $m_1m_2$ vertex in second layer is defined as $\frac{\alpha_{m_1}} {S_{m_1m_2}+0.5}$, which varies proportionally with $\alpha_{m_1}$ but grows in inverse proportion to $S_{m_1m_2}$, so are the other weights of edges. Then, the weight for $m_1m_2$ vertex is defined as:
\begin{equation}
\centering
   \hat{\alpha}_{m_1m_2}=\frac{\alpha_{m_1}+\alpha_{m_2}}{S_{m_1m_2}+0.5}
\end{equation}

\begin{equation}
\centering
   \alpha_{m_1m_2}= \frac{e^{\hat{\alpha}_{m_1m_2}}}{\sum_{m_i \ne m_j} e^{\hat{\alpha}_{m_im_j}}}
\end{equation}
where $\alpha_{m_1m_2}$ is a scalar that represents the weight of $m_1m_2$ vertex (the value 0.5 on the denominator is applied to control the relative importance of similarity and vertex weights to $\alpha_{m_1m_2}$), and Eq.(14) is a softmax-uniformed operation of weight vector in second layer. The output of the second layer is the weighted average of information in bimodal vertices:
\begin{equation}
\centering
   \bm{B}= \sum \alpha_{m_1m_2} \cdot \bm{V}_{m_1m_2}
\end{equation}
where $\bm{B}$ is the combined bimodal dynamic.

Each two bimodal vertices are further fused using the same $MLP$ network that generate bimodal dynamics (but the parameters are not shared) to obtain trimodal vertices in the third layer, i.e., trimodal dynamic learning layer. In addition, as shown in Fig.~\ref{8}, each specific bimodal vertex is also fused with the unimodal vertex that do not contribute to the formation of this bimodal vertex in previous fusion stage, resulting in three other trimodal vertices. Therefore, there are six trimodal vertices in total. We apply the same weight computing method for edges that link to the third layers and the same importance measure method for trimodal vertices as applied in second layer. Then we add up the weighted information from each trimodal vertex to obtain final trimodal information $\bm{T}$ .

The final output of the hierarchical graph fusion network is the concatenation of trimodal, bimodal and unimodal dynamics, defined as: $\bm{\Omega}=\bm{U} \oplus \bm{B} \oplus \bm{T}$. Finally, a decision neural network is utilized to infer the final decision:
\begin{equation}
\centering
   \textbf{M}=\emph{Dec}(\bm{\Omega};\Theta_{Dec})
\end{equation}
where $\textbf{M} \in \mathbb{R}^{N}$ ($N$ is the number of classes). $\emph{Dec}:$ $\mathbb{R}^{3k}\rightarrow \mathbb{R}^{N}$  is the decision inference module containing a normalization layer followed by three dense layers activated by $tanh$, $tanh$ and $softmax$ respectively.

\section{Experiments}\label{sec:Experiments}

\subsection{Datasets}
$\textbf{CMU-MOSI}$ \cite{Zadeh2016Multimodal} is a widely-used dataset that includes a collection of 93 opinion videos from different speakers which have been divided into 2199 segments in total. We report the binary (positive and negative) sentiment accuracy and F1 score on this dataset. We use 1141 segments for training, 306 segments for validation and 752 segments for testing. $\textbf{CMU-MOSEI}$ \cite{MOSEI}, consisting of 2928 videos (20802 segments in total), is the largest multimodal language analysis dataset so far. We report positive, negative and neutral sentiment classification accuracy and F1 score on this dataset. We use 13168 segments for training, 3020 segments for validation and 4614 segments for testing. $\textbf{IEMOCAP}$ \cite{Busso2008IEMOCAP} is an emotion recognition dataset that contains 151 videos and the videos have been segmented into 7433 segments. The dataset contains 9 emotional labels. We analyze the anger, happiness, sadness and neutral emotions so as to compare with previous approaches. The training set consists of 4674 segments, while the validation and test set has 1136 and 1623 segments, respectively.

\begin{figure*}
\centering
\includegraphics[width=1.9\columnwidth]{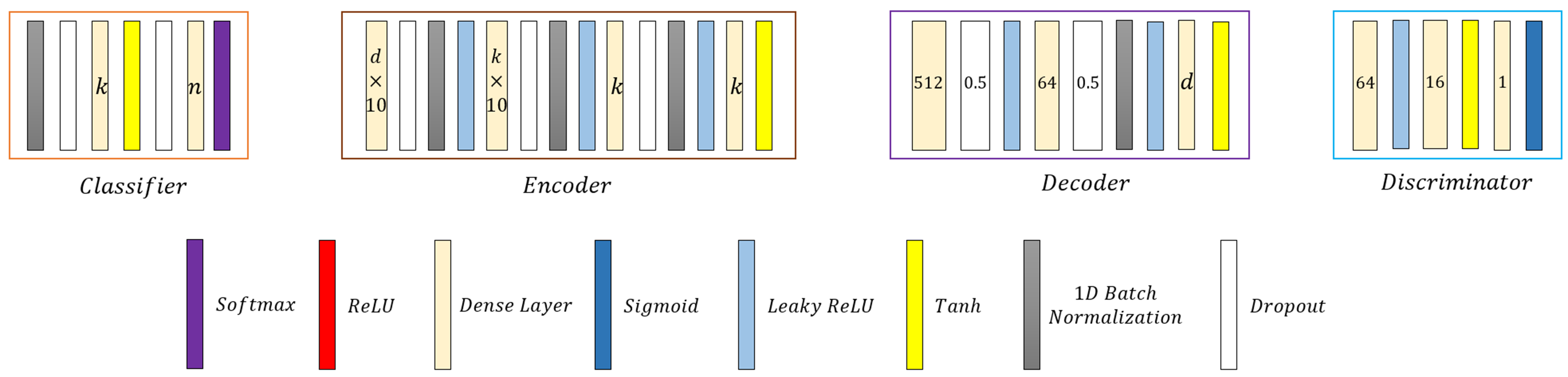}
\caption{\label{888}Schematic Diagram of Encoder, Decoder, Discriminator and Classifier.
}
\end{figure*}

\subsection{Experimental details}

We develop our model on Pytorch. The implemental details of encoder, decoder, classifier and discriminator are shown in Fig.~\ref{888}. Specifically, the number inside the dense layer is the output dimensionality, $k$ is the dimensionality of encoded representations, $n$ is the number of classes and $d$ denotes the dimensionality of input feature vector ($d$ is 50 for CMU-MOSI and CMU-MOSEI datasets and 100 for IEMOCAP datatset). We apply Mean Square Error as loss function for graph fusion network with Adam \cite{Kingma2014Adam} being the optimizer. The framework is trained end to end. The hyper-parameters such as batch size, learning rate and $k$ are chosen by grid search to optimize the performance.

In feature extraction stage, for CMU-MOSI and IEMOCAP, we follow the setting of \cite{Poria2017Context}. A text-CNN consisting of word2vec embedding \cite{Mikolov2013Efficient} followed by CNNs \cite{Karpathy2014Large}, openSMILE \cite{Eyben2010Opensmile} and 3D-CNN \cite{Ji20133D} are applied for language, acoustic and visual feature extraction respectively.
For CMU-MOSEI, the features are extracted using CMU-MultimodalSDK \footnote{https://github.com/A2Zadeh/CMU-MultimodalSDK}.  GloVe word embeddings \cite{pennington2014glove}, Facet  \footnote{ iMotions 2017. https://imotions.com/} and COVAREP \cite{Degottex2014COVAREP} are applied for extracting language, visual and acoustic features respectively (please refer to \cite{HFFN} for more details).

After extraction, similar to HFFN \cite{HFFN}, we develop a Unimodal Context Learning Network (UCLN): $\mathbb{R}^{u\times d_j}\! \rightarrow \! \mathbb{R}^{u\times d} $, which is composed of a bidirectional LSTM layer followed by a dense layer, for each separate modality. Here, $u$ denotes the number of segments that constitute a video and $d_j$ is the dimensionality of raw feature vector for $j^{th}$ modality. Through UCLN, feature vectors of all modalities are mapped into the same dimensionality $d$ ($d_a=d_v=d_l=d$). UCLN for each modality is individually trained followed by a dense layer: $\mathbb{R}^{d}\!\rightarrow\! \mathbb{R}^{N}$ using  Adadelta  \cite{ADADELTA} as optimizer and  categorical cross-entropy as loss function. The precessed feature vectors output by UCLN will be sent into ARGF for modality fusion.

\subsection{Results and Discussions}

\begin{table}[t]
\centering
 \resizebox{.95\columnwidth}{!}{\begin{tabular}{c|c|c}
 \hline
   Methods & Avg Accuracy    & Avg F1 score\\
 \hline
  LMFN \cite{LMFN}  & 80.9  &  80.9 \\
   TFN  \cite{Zadeh2017Tensor}  & 78.3  &  78.3 \\
 LMF  \cite{Liu2018Efficient} & 75.8 & 75.9  \\
  CHF-Fusion \cite{Context}& 80.0 &  - \\
  BC-LSTM \cite{Poria2017Context} & 77.9  &  78.1 \\
 HFFN \cite{HFFN} & 80.2 & 80.3 \\
 \hline
 ARGF ($a,v\rightarrow l$)& \textbf{81.38}  &  \textbf{81.52} \\
   ARGF ($l,v\rightarrow a$)& 81.25 &  81.32 \\
  ARGF ($a,l\rightarrow v$)& 81.12  &  81.25 \\
 \hline
 \end{tabular}}
  \caption{ \label{t4}Performance on CMU-MOSI dataset.}
\end{table}%

\begin{table*}[t]
\centering
 \resizebox{1.9\columnwidth}{!}{\begin{tabular}{c|c|c|c|c|c|c|c|c}
 \hline
 &\multicolumn{8}{c}{IEMOCAP }\\
 \hline
  Models &\ \ Happy\ \ &\ \ \ \ Sad\ \ \ \ &\ Neutral\ &\ \ Angry\ \ &\ Excited\ & Frustrated & Avg Accuracy & Avg F1 score\\
 \hline
 LMFN \cite{LMFN} & 31.25 & 64.90 & 58.33 & 62.94 &  47.83  & 69.55 &  58.10  &  57.88\\
CHFusion \cite{Context} & \textbf{36.11} & 62.04 & 56.25 & \textbf{70.00} &  55.52  &  63.52 &  58.35  &  58.39\\
 BC-LSTM  \cite{Poria2017Context} & 35.42 & 58.37 & 53.91 & 64.71 &  54.18  &  60.63 &  55.70  &  55.78\\
TFN \cite{Zadeh2017Tensor}  & 29.86 & 55.51 & 48.81 & 60.59 &  57.86  &  63.25 &  54.28  &  54.19\\
LMF \cite{Liu2018Efficient}  & 26.39 & 49.39 & 56.77 & 61.18 &  47.16  &  63.25 &  53.17  &  53.02\\
HFFN \cite{HFFN} & 31.25 & 63.67 & 54.69 & 61.18 &  48.83  &  \textbf{69.82} &  57.12  &  56.82\\
 \hline
   ARGF ($a,v\rightarrow l$)&15.28 & \textbf{68.98} & \textbf{59.11}&68.24 &\textbf{72.58}& 61.42 & \textbf{60.69} & 59.53\\
 ARGF ($l,v\rightarrow a$)& 22.92 &  62.86 & 57.55 & 65.29 &  67.22 &  67.45 & 60.20 & 59.75\\
  ARGF ($l,a\rightarrow v$)& 26.39 &  \textbf{68.98} & 54.95 & 62.35 &  64.21 &  68.50 & 60.20 & \textbf{59.81}\\
 \hline
 \end{tabular}}
  \caption{ \label{t5}Performance on IEMOCAP dataset. The evaluation index for each emotion is the recognition accuracy.}
\end{table*}%

\begin{table*}[t]
\centering
 \resizebox{1.9\columnwidth}{!}{\begin{tabular}{c|c|c|c|c|c|c|c|c}
 \hline
   & \multicolumn{2}{c|}{Positive} & \multicolumn{2}{c|}{Negative} & \multicolumn{2}{c|}{Neutral} & \multicolumn{2}{c}{Average} \\
 \hline
 Models  & Accuracy & F1 score & Accuracy &  F1 score & Accuracy & F1 score & Avg Accuracy & Avg F1 score\\
 \hline
 LMFN \cite{LMFN} & 61.88 & 59.98 & 26.21 & 31.42 & 75.54 & 71.49 & 60.77 & 59.42 \\
 TFN \cite{Zadeh2017Tensor} & 60.46 & 58.01 & 18.30 & 25.08 & 77.70 & 70.91 &  59.69 & 57.13\\
 LMF \cite{Liu2018Efficient} & \textbf{68.91} & 59.85 & 11.77 &  18.31 & 75.06 & 70.15 &  59.41 & 55.80\\
 CHFusion  \cite{Context}& 59.19 & 56.61 & 22.55 & 27.96 & 73.69 & 69.56 &  58.28 &   56.69\\
 BC-LSTM  \cite{Poria2017Context}& 64.20 & \textbf{60.97} & 24.53 & 29.74 & 74.53 & 71.19 &  60.58 & 59.14\\
  HFFN \cite{HFFN} & 59.49 & 59.05 & 26.61 & 31.35 & 75.85 & 71.35 & 60.32 & 59.03 \\
 \hline
  ARGF ($a,v\rightarrow l$) & 64.57 & 60.55 & 21.66 & 28.01 & 76.20 & 71.77 & \textbf{60.88}  &  58.92 \\
  ARGF ($l,v\rightarrow a$) & 61.29 & 59.61 & \textbf{28.78} & \textbf{33.31} & 74.30 & 71.16 &60.55  &  \textbf{59.52} \\
  ARGF ($a,l\rightarrow v$) & 62.33 & 60.04 & 20.87 & 26.99 & \textbf{77.88} & \textbf{71.99} & \textbf{60.88}  &  58.66 \\
  \hline
 \end{tabular}}
  \caption{ \label{t6}Performance on CMU-MOSEI dataset.}
\end{table*}%

\textbf{Comparison with Baselines:}\label{sec:multimodal}
As presented in Table~\ref{t4}, ARGF shows improvement over typical approaches and  outperforms the SOTA method LMFN by about 0.5\% on CMU-MOSI dataset. Moreover, compared to the tensor fusion methods TFN and LMF, ARGF achieves improvement by about 3\% and 6\% respectively. We argue that it is probably partly because these approaches ignore exploring modality-invariant embedding space, while we adopt adversarial training to learn joint embedding space before fusion. These results demonstrate the superiority of ARGF. We also report ARGF's performance on the more challenging datasets IEMOCAP and CMU-MOSEI to evaluate ARGF's robustness. For IEMOCAP, from Table~\ref{t5} we can infer that ARGF achieves the best performance and significantly outperforms SOTA methods by about 2\% in the most important index (average accuracy). For CMU-MOSEI, as shown in Table~\ref{t6}, the average accuracy and F1 score of ARGF are still the highest among all methods, showing excellent performance.  In addition, all the models' performance on `Negative' emotion are weaker than that on other emotions by a large margin. We speculate that one of the reasons is that the number of samples belonging to `Negative' class is much fewer than the ones of other emotions. Consequently the model tends to predict otherwise which is less `risky'.

\begin{table}[t]
\centering
 \resizebox{.95\columnwidth}{!}{\begin{tabular}{c|c|c}
 \hline
   Methods  & Number of Parameters & FLOPs\\
 \hline
  BC-LSTM \cite{Poria2017Context}  & 1,383,902 & \textbf{1,322,044} \\
 TFN \cite{Zadeh2017Tensor}  & 4,245,986 & 8,491,844\\
 \hline
 ARGF (ours)  & \textbf{1,270,770} & 2,017,645 \\
 \hline
 \end{tabular}}
  \caption{ \label{t222}The Comparison of Model Complexity.}
\end{table}%

\textbf{Complexity Analysis:}
We use the amount of trainable parameters as a proxy for the space complexity. As shown in Table~\ref{t222}, our model has 1,270,770 trainable parameters ($k$ is set to 50), which is approximately 29.93\% and 91.83\% of the number of parameters of TFN and BC-LSTM, respectively. To explore the time complexity of ARGF and make a comparison against baselines, we compute FLOPs for each method during testing. We empirically find that ARGF needs 2,017,645 FLOPs in testing, whereas the number of FLOPs in TFN and BC-LSTM are 8,491,844 and 1,322,044, respectively. The reason for fewer trainable parameters and moderate number of FLOPs is that despite having multiple components in the architecture, our model entails low dimensions for unimodal and multimodal representations. In addition, every component in our architecture has only several layers, which guarantees a reasonable computational load.

\textbf{Choice of Target Modality:}\label{sec:target} As presented in Table~\ref{t4}, ~\ref{t5} and ~\ref{t6}, the results demonstrate that language modality is the best choice for CMU-MOSI, while acoustic and visual arrives in the second and third place respectively by a slight margin. Actually, all the modalities perform very closely across the three datasets. There are no hard and fast rules to choose an optimal target modality since their performances are rather similar. It is reasonable because the target modality only serves as a distribution provider so that the source modalities can map their transformed distributions into that of the target modality. During the distribution mapping procedure, all the modalities' information is well preserved by the use of decoders and classifier. Therefore, theoretically there is no difference what the target modality is.

\begin{table}[t]
\centering
 \resizebox{.95\columnwidth}{!}{\begin{tabular}{c|c|c}
 \hline
    &\ Accuracy\ &\   F1 score\ \\
 \hline
 ARGF (Without Adversarial Training) & 80.11 & 80.23  \\
 ARGF (Without Classifier) & 80.92 & 80.98  \\
 ARGF (Without Decoder) & 79.72 & 79.76 \\
 \hline
  ARGF & \textbf{81.38} & \textbf{81.52}  \\
 \hline
 \end{tabular}}
  \caption{ \label{t8}Comparison of Different Architectures on CMU-MOSI dataset.}
\end{table}

\textbf{Ablation Studies:} From Table~\ref{t8} we can see that the removal of any component in ARGF leads to a decline in performance. Specifically, the removal of decoders leads to the most significant decline of more than 1.5\%, closely followed by the removal of adversarial training. In addition, experiments without classifier achieve relatively acceptable results, with a slight decrease of around 0.4\%, but classifier is critical for learning a discriminative embedding space as demonstrated in the visualization part. The ablation studies show that these three components are crucial factors contributing to the competitive results of ARGF.

\begin{table}[t]
\centering
 \resizebox{.95\columnwidth}{!}{\begin{tabular}{c|c|c}
 \hline
  &\ CMU-MOSI\ &
  \ IEMOCAP\ \\
 \hline
  Concatenation + FC & 80.19  &  57.98  \\
  Multiplication + FC & 80.05  &  58.60 \\
   Weighted Average & 78.86  &  59.64  \\
 Tensor Fusion & 79.65  &  59.46  \\
 Low-rank Modality Fusion (LMF) & 78.72  &  59.21  \\
 Dynamic Fusion Graph (DFG) & 79.92  &  57.79 \\
 \hline
 Graph Fusion Network (GFN, ours) & \textbf{81.38}  &  \textbf{60.20} \\
 \hline
 \end{tabular}}
  \caption{ \label{t7}Accuracy of various fusion structures. The diagrams of some fusion methods for comparison are given in Fig.~\ref{9}. Please refer to the corresponding papers \cite{Zadeh2017Tensor,Liu2018Efficient,MOSEI} for the diagrams of Tensor Fusion, LMF and DFG respectively.}
\end{table}%

\begin{figure}
\centering
\includegraphics[width=.95\columnwidth]{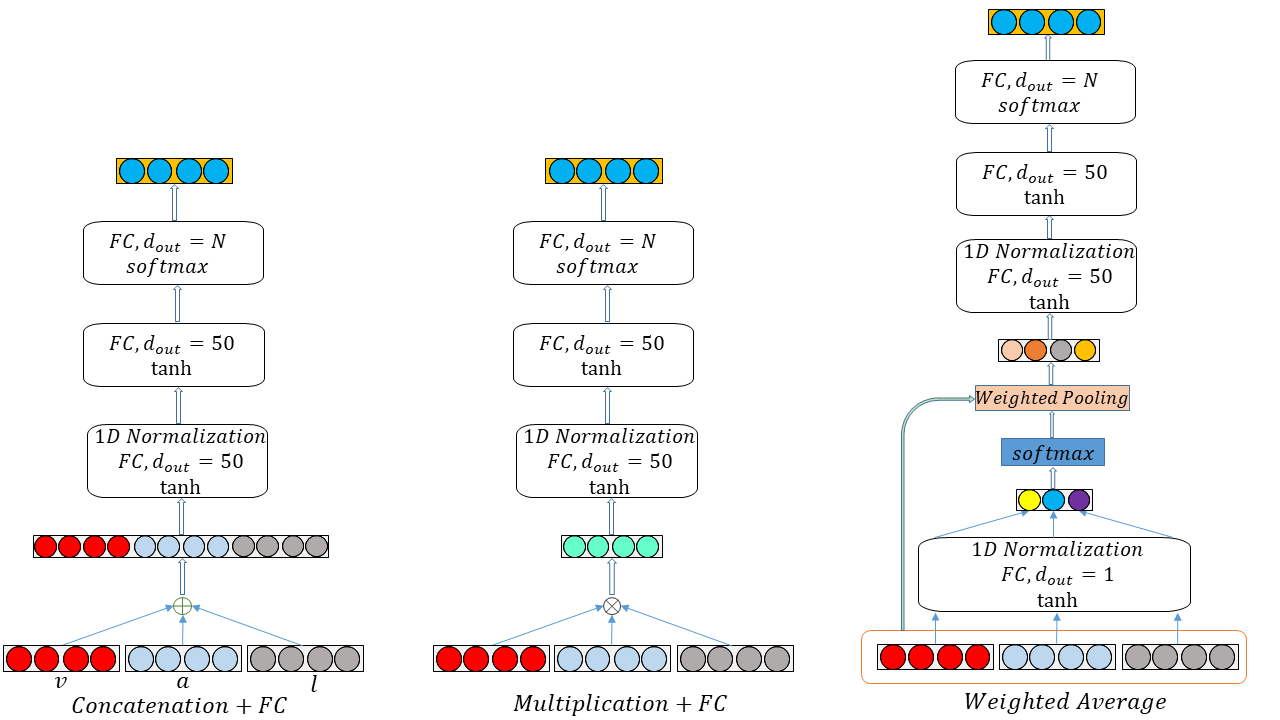}
\caption{\label{9}Schematic Diagram of `Concatenation + FC', `Multiplication + FC' and `Weighted Average'.
}
\end{figure}

\textbf{Comparison between fusion strategies:} To demonstrate that our fusion method is indeed effective, we conduct a contrast experiment to compare the performance between other fusion strategies with our GFN.
We can infer from Table~\ref{t7} that GFN brings significant improvement on performance compared to other fusion methods, demonstrating its superiority. Specifically, DFG \cite{MOSEI} achieves comparable results in two datasets, but our GFN outperforms it by over 2\% in IEMOCAP and 1\% in CMU-MOSI. We argue that it is because we explicitly model unimodal, bimodal and trimodal dynamics and obtain trimodal representations in a more comprehensive way (as also revealed in the following visualization part). From above analysis, we can draw the conclusion that the interpretable, comprehensive and hierarchical fusion strategy is a crucial factor that lead to the marked improvement of ARGF. In addition, all the fusion strategies fitted in our framework obtain relatively good results compared with the original frameworks. For instance, our version of Tensor Fusion \cite{Zadeh2017Tensor} achieves the accuracy of 59.46\% in IEMOCAP, which is much higher than TFN's accuracy (54.28\%, see Table~\ref{t5}). To some extent, these comparisons prove that learning joint embedding space before feature fusion is indeed important and effective.

\begin{figure}
\centering
\includegraphics[width=.95\columnwidth]{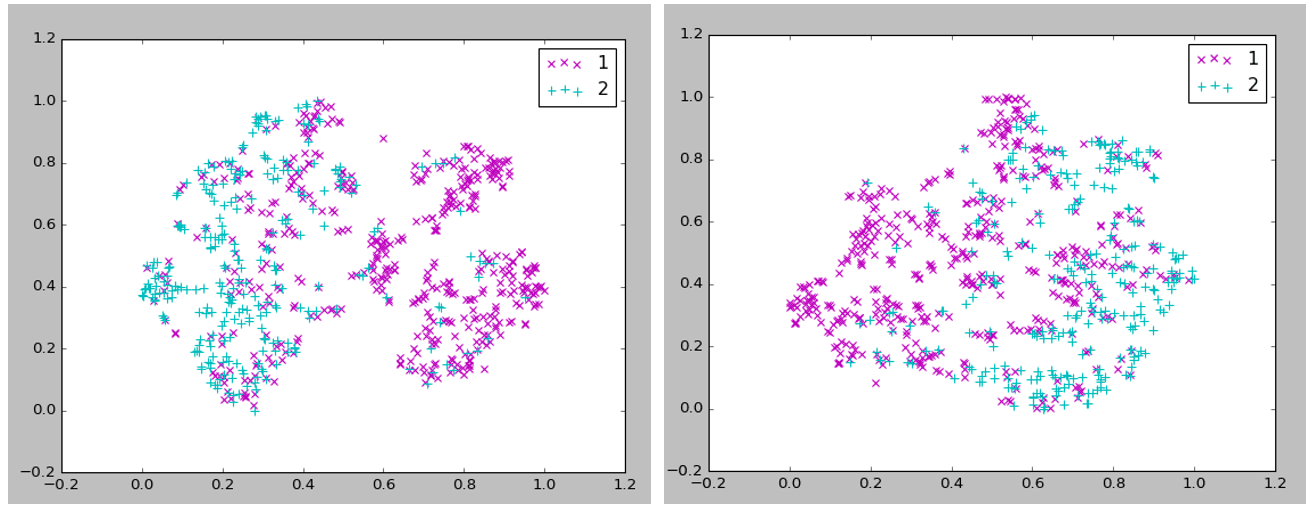}
\caption{\label{5}Visualization for distributions of different sentiments in learnt embedding space. The cyan and purple dots represent positive and negative sentiment respectively.
}
\end{figure}

\textbf{Visualization of Embedding Space:}\label{sec:visual} We provide a visualization for distributions of sentiments in the embedding space where the the left sub-figure on Fig.~\ref{5} illustrates the embedding space learnt by ARGF while the right sub-figure presents the embedding space learnt without classifier. The visualization is obtained by transforming the concatenated features of all modalities output by encoders. We use t-SNE algorithm to transform high dimensional concatenated feature vectors into 2-dimension feature points.  We can infer from Fig.~\ref{5} that in the embedding space learnt by ARGF, the dots tend to form two separate clusters which mainly belong to positive and negative sentiment respectively. The distance between two clusters is large but the dots belonging to same cluster are tied closely, which demonstrates the discrimination of our embedding space. Nevertheless, there are some dots that are extremely difficult to be correctly classified found in the wrong clusters, which drives the need for advanced fusion strategies to explore cross-modal dynamics. In contrast, it is clear that the embedding space learnt without classifier cannot distinguish positive and negative sentiments effectively, which highlights the necessity of classifier in learning a discriminative embedding space.

\begin{figure}
\centering
\includegraphics[width=.95\columnwidth]{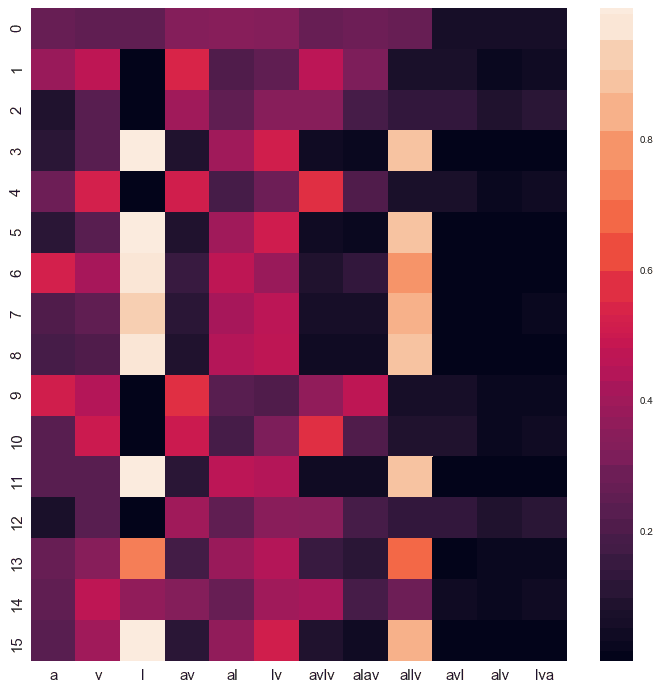}
\caption{\label{6}Visualization for Graph Fusion in CMU-MOSI sentiment analysis task. Each row represents one test sample and each column denotes weight for one vertex.
}
\end{figure}

\textbf{Visualization for Weights of Vertices in Graph Fusion:}\label{sec:graph} This visualization is conducted to prove graph fusion's interpretability. As shown in Fig.~\ref{6}, for unimodal interactions (the first three columns), obviously language modality is the most predictive for the majority of samples, which is reasonable since language is the most important clue for sentiment analysis. But its importance varies greatly across samples, indicating the advantage of analyzing sentiment in a multimodal form for other modalities can play a dominant role whenever language modality is trivial for a specific sample. For bimodal interactions (from $4^{th}$ to $6^{th}$ columns), weights for $al$ vertex ($\alpha_{al}$) and $vl$ vertex ($\alpha_{vl}$) are very close, followed closely by $\alpha_{av}$, possibly because language modality plays a huge role in bimodal fusion. For trimodal interactions, to our surprise, vertices obtained by fusing a bimodal vertex and a unimodal vertex hardly make any difference, but vertices obtained by fusing two bimodal vertices dominate trimodal information. It proves the necessity to explore interactions for each two bimodal vertices, which are not modeled by DFG \cite{MOSEI}.

\section{Conclusions}

We propose a novel multimodal framework to learn a discriminative joint embedding space and then perform graph fusion. By introducing adversarial training to match distributions, modality gap can be significantly narrowed and the representations can be directly fused. With the aid of GFN, we can explore unimodal, bimodal and trimodal dynamics successively and dynamically change the fusion structure.

\section{Acknowledgements}
This work is supported by the National Key R\&D Program of China under Grant No. 2018YFB1601101 of No. 2018YFB1601100.

%\begin{quote}
%\begin{small}
%\textbackslash bibliographystyle\{aaai\}
%\textbackslash bibliography{sentiment2}
%\end{small}
%\end{quote}
\small
\bibliographystyle{aaai}
\bibliography{sentiment2}

\begin{thebibliography}{}

\bibitem[\protect\citeauthoryear{Baltru\v{s}aitis, Ahuja, and
  Morency}{2019}]{8269806}
Baltru\v{s}aitis, T.; Ahuja, C.; and Morency, L.-P.
\newblock 2019.
\newblock Multimodal machine learning: A survey and taxonomy.
\newblock {\em IEEE Transactions on Pattern Analysis and Machine Intelligence}
  41(2):423--443.

\bibitem[\protect\citeauthoryear{Busso \bgroup et al\mbox.\egroup
  }{2008}]{Busso2008IEMOCAP}
Busso, C.; Bulut, M.; Lee, C.~C.; Kazemzadeh, A.; Mower, E.; Kim, S.; Chang,
  J.~N.; Lee, S.; and Narayanan, S.~S.
\newblock 2008.
\newblock Iemocap: interactive emotional dyadic motion capture database.
\newblock {\em Language Resources and Evaluation} 42(4):335--359.

\bibitem[\protect\citeauthoryear{Degottex \bgroup et al\mbox.\egroup
  }{2014}]{Degottex2014COVAREP}
Degottex, G.; Kane, J.; Drugman, T.; Raitio, T.; and Scherer, S.
\newblock 2014.
\newblock Covarep: A collaborative voice analysis repository for speech
  technologies.
\newblock In {\em ICASSP},  960--964.

\bibitem[\protect\citeauthoryear{Eyben}{2010}]{Eyben2010Opensmile}
Eyben, F.
\newblock 2010.
\newblock Opensmile: the munich versatile and fast open-source audio feature
  extractor.
\newblock In {\em ACM International Conference on Multimedia},  1459--1462.

\bibitem[\protect\citeauthoryear{Garcia and Bruna}{2018}]{graph}
Garcia, V., and Bruna, J.
\newblock 2018.
\newblock Few-shot learning with graph neural networks.
\newblock In {\em ICLR}.

\bibitem[\protect\citeauthoryear{Goodfellow \bgroup et al\mbox.\egroup
  }{2014}]{GAN}
Goodfellow, I.~J.; Pouget-Abadie, J.; Mirza, M.; Bing, X.; Warde-Farley, D.;
  Ozair, S.; Courville, A.; and Bengio, Y.
\newblock 2014.
\newblock Generative adversarial nets.
\newblock In {\em NeurIPS}.

\bibitem[\protect\citeauthoryear{Gu \bgroup et al\mbox.\egroup
  }{2018}]{Gu2018Multimodal}
Gu, Y.; Yang, K.; Fu, S.; Chen, S.; Li, X.; and Marsic, I.
\newblock 2018.
\newblock Multimodal affective analysis using hierarchical attention strategy
  with word-level alignment.
\newblock In {\em ACL},  2225--2235.

\bibitem[\protect\citeauthoryear{Ji, Yang, and Yu}{2013}]{Ji20133D}
Ji, S.; Yang, M.; and Yu, K.
\newblock 2013.
\newblock 3d convolutional neural networks for human action recognition.
\newblock {\em IEEE Transactions on Pattern Analysis and Machine Intelligence}
  35(1):221--231.

\bibitem[\protect\citeauthoryear{Kampman \bgroup et al\mbox.\egroup
  }{2018}]{Personality}
Kampman, O.; Barezi, E.~J.; Bertero, D.; and Fung, P.
\newblock 2018.
\newblock Investigating audio, visual, and text fusion methods for end-to-end
  automatic personality prediction.
\newblock In {\em ACL short paper}.

\bibitem[\protect\citeauthoryear{Karpathy \bgroup et al\mbox.\egroup
  }{2014}]{Karpathy2014Large}
Karpathy, A.; Toderici, G.; Shetty, S.; Leung, T.; Sukthankar, R.; and Li,
  F.~F.
\newblock 2014.
\newblock Large-scale video classification with convolutional neural networks.
\newblock In {\em CVPR},  1725--1732.

\bibitem[\protect\citeauthoryear{Kingma and Ba}{2015}]{Kingma2014Adam}
Kingma, D.~P., and Ba, J.
\newblock 2015.
\newblock Adam: A method for stochastic optimization.
\newblock In {\em ICLR}.

\bibitem[\protect\citeauthoryear{Liang \bgroup et al\mbox.\egroup
  }{2018}]{RMFN}
Liang, P.~P.; Liu, Z.; Zadeh, A.; and Morency, L.~P.
\newblock 2018.
\newblock Multimodal language analysis with recurrent multistage fusion.
\newblock In {\em EMNLP},  150--161.

\bibitem[\protect\citeauthoryear{Liang \bgroup et al\mbox.\egroup }{2019}]{MMB}
Liang, P.~P.; Lim, Y.~C.; Tsai, Y.~H.; Salakhutdinov, R.~R.; and Morency, L.-P.
\newblock 2019.
\newblock Strong and simple baselines for multimodal utterance embeddings.
\newblock In {\em NAACL},  2599--2609.

\bibitem[\protect\citeauthoryear{Lin \bgroup et al\mbox.\egroup
  }{2015}]{Lin2015Semantics}
Lin, Z.; Ding, G.; Hu, M.; and Wang, J.
\newblock 2015.
\newblock Semantics-preserving hashing for cross-view retrieval.
\newblock In {\em CVPR},  3864--3872.

\bibitem[\protect\citeauthoryear{Liu and Zhang}{2012}]{Liu2012}
Liu, B., and Zhang, L.
\newblock 2012.
\newblock {\em A Survey of Opinion Mining and Sentiment Analysis}.
\newblock Boston, MA: Springer US.
\newblock  415--463.

\bibitem[\protect\citeauthoryear{Liu \bgroup et al\mbox.\egroup
  }{2018}]{Liu2018Efficient}
Liu, Z.; Shen, Y.; Liang, P.~P.; Zadeh, A.; and Morency, L.~P.
\newblock 2018.
\newblock Efficient low-rank multimodal fusion with modality-specific factors.
\newblock In {\em ACL},  2247--2256.

\bibitem[\protect\citeauthoryear{Mai, Hu, and Xing}{2019}]{HFFN}
Mai, S.; Hu, H.; and Xing, S.
\newblock 2019.
\newblock Divide, conquer and combine: Hierarchical feature fusion network with
  local and global perspectives for multimodal affective computing.
\newblock In {\em ACL},  481--492.
\newblock Association for Computational Linguistics.

\bibitem[\protect\citeauthoryear{{Mai}, {Xing}, and {Hu}}{2019}]{LMFN}
{Mai}, S.; {Xing}, S.; and {Hu}, H.
\newblock 2019.
\newblock Locally confined modality fusion network with a global perspective
  for multimodal human affective computing.
\newblock {\em IEEE Transactions on Multimedia}  1--1.

\bibitem[\protect\citeauthoryear{Majumder \bgroup et al\mbox.\egroup
  }{2018}]{Context}
Majumder, N.; Hazarika, D.; Gelbukh, A.; Cambria, E.; and Poria, S.
\newblock 2018.
\newblock Multimodal sentiment analysis using hierarchical fusion with context
  modeling.
\newblock {\em Knowledge-based systems} 161:124--133.

\bibitem[\protect\citeauthoryear{Makhzani \bgroup et al\mbox.\egroup
  }{2016}]{AAE}
Makhzani, A.; Shlens, J.; Jaitly, N.; and Goodfellow, I.~J.
\newblock 2016.
\newblock Adversarial autoencoders.
\newblock In {\em ICLR Workshop}.

\bibitem[\protect\citeauthoryear{Mikolov \bgroup et al\mbox.\egroup
  }{2013}]{Mikolov2013Efficient}
Mikolov, T.; Chen, K.; Corrado, G.; and Dean, J.
\newblock 2013.
\newblock Efficient estimation of word representations in vector space.
\newblock In {\em ICLR Workshop},  1725--1732.

\bibitem[\protect\citeauthoryear{Nojavanasghari \bgroup et al\mbox.\egroup
  }{2016}]{Nojavanasghari2016Deep}
Nojavanasghari, B.; Gopinath, D.; Koushik, J.; and Morency, L.~P.
\newblock 2016.
\newblock Deep multimodal fusion for persuasiveness prediction.
\newblock In {\em Proceedings of ACM International Conference on Multimodal
  Interaction},  284--288.

\bibitem[\protect\citeauthoryear{Peng and Qi}{2019}]{CM-GAN}
Peng, Y., and Qi, J.
\newblock 2019.
\newblock Cm-gans: cross-modal generative adversarial networks for common
  representation learning.
\newblock {\em ACM Transactions on Multimedia Computing, Communications, and
  Applications (TOMM)} 15(1):22.

\bibitem[\protect\citeauthoryear{Pennington, Socher, and
  Manning}{2014}]{pennington2014glove}
Pennington, J.; Socher, R.; and Manning, C.~D.
\newblock 2014.
\newblock Glove: Global vectors for word representation.
\newblock In {\em Empirical Methods in Natural Language Processing (EMNLP)},
  1532--1543.

\bibitem[\protect\citeauthoryear{Pham \bgroup et al\mbox.\egroup }{2019}]{MCTN}
Pham, H.; Liang, P.~P.; Manzini, T.; Morency, L.~P.; and Barnab\v{a}s, P.
\newblock 2019.
\newblock Found in translation: Learning robust joint representations by cyclic
  translations between modalities.
\newblock In {\em AAAI},  6892--6899.

\bibitem[\protect\citeauthoryear{Poria \bgroup et al\mbox.\egroup
  }{2016}]{Poria2017Convolutional}
Poria, S.; Chaturvedi, I.; Cambria, E.; and Hussain, A.
\newblock 2016.
\newblock Convolutional mkl based multimodal emotion recognition and sentiment
  analysis.
\newblock In {\em ICDM},  439--448.

\bibitem[\protect\citeauthoryear{Poria \bgroup et al\mbox.\egroup
  }{2017a}]{Poria2017A}
Poria, S.; Cambria, E.; Bajpai, R.; and Hussain, A.
\newblock 2017a.
\newblock A review of affective computing: From unimodal analysis to multimodal
  fusion.
\newblock {\em Information Fusion} 37:98--125.

\bibitem[\protect\citeauthoryear{Poria \bgroup et al\mbox.\egroup
  }{2017b}]{Poria2017Context}
Poria, S.; Cambria, E.; Hazarika, D.; Majumder, N.; Zadeh, A.; and Morency,
  L.~P.
\newblock 2017b.
\newblock Context-dependent sentiment analysis in user-generated videos.
\newblock In {\em ACL},  873--883.

\bibitem[\protect\citeauthoryear{Su, Lan, and Liu}{2018}]{Su2018Multimodal}
Su, R.; Lan, W.; and Liu, X.
\newblock 2018.
\newblock Multimodal learning using 3d audio-visual data for audio-visual
  speech recognition.
\newblock In {\em International Conference on Asian Language Processing}.

\bibitem[\protect\citeauthoryear{Tsai \bgroup et al\mbox.\egroup
  }{2019a}]{MULT}
Tsai, Y.-H.~H.; Bai, S.; Liang, P.~P.; Kolter, J.~Z.; Morency, L.-P.; and
  Salakhutdinov, R.
\newblock 2019a.
\newblock Multimodal transformer for unaligned multimodal language sequences.
\newblock In {\em ACL},  6558--6569.

\bibitem[\protect\citeauthoryear{Tsai \bgroup et al\mbox.\egroup }{2019b}]{MFM}
Tsai, Y. H.~H.; Liang, P.~P.; Zadeh, A.; Morency, L.~P.; and Salakhutdinov, R.
\newblock 2019b.
\newblock Learning factorized multimodal representations.
\newblock In {\em ICLR}.

\bibitem[\protect\citeauthoryear{Wang \bgroup et al\mbox.\egroup
  }{2019}]{RAVEN}
Wang, Y.; Shen, Y.; Liu, Z.; Liang, P.~P.; and Zadeh, A.
\newblock 2019.
\newblock Words can shift: Dynamically adjusting word representations using
  nonverbal behaviors.
\newblock In {\em AAAI},  7216--7223.

\bibitem[\protect\citeauthoryear{Wollmer \bgroup et al\mbox.\egroup
  }{2013}]{Wollmer2013YouTube}
Wollmer, M.; Weninger, F.; Knaup, T.; Schuller, B.; Sun, C.; Sagae, K.; and
  Morency, L.~P.
\newblock 2013.
\newblock Youtube movie reviews: Sentiment analysis in an audio-visual context.
\newblock {\em IEEE Intelligent Systems} 28(3):46--53.

\bibitem[\protect\citeauthoryear{Xing \bgroup et al\mbox.\egroup
  }{2018}]{Xing2018Deep}
Xing, X.; Li, H.; Lu, H.; Gao, L.; and Ji, Y.
\newblock 2018.
\newblock Deep adversarial metric learning for cross-modal retrieval.
\newblock In {\em World Wide Web},  1--16.

\bibitem[\protect\citeauthoryear{Zadeh \bgroup et al\mbox.\egroup
  }{2016}]{Zadeh2016Multimodal}
Zadeh, A.; Zellers, R.; Pincus, E.; and Morency, L.~P.
\newblock 2016.
\newblock Multimodal sentiment intensity analysis in videos: Facial gestures
  and verbal messages.
\newblock {\em IEEE Intelligent Systems} 31(6):82--88.

\bibitem[\protect\citeauthoryear{Zadeh \bgroup et al\mbox.\egroup
  }{2017}]{Zadeh2017Tensor}
Zadeh, A.; Chen, M.; Poria, S.; Cambria, E.; and Morency, L.~P.
\newblock 2017.
\newblock Tensor fusion network for multimodal sentiment analysis.
\newblock In {\em EMNLP},  1114--1125.

\bibitem[\protect\citeauthoryear{Zadeh \bgroup et al\mbox.\egroup
  }{2018a}]{Zadeh2018Memory}
Zadeh, A.; Liang, P.~P.; Mazumder, N.; Poria, S.; Cambria, E.; and Morency,
  L.~P.
\newblock 2018a.
\newblock Memory fusion network for multi-view sequential learning.
\newblock In {\em AAAI},  5634--5641.

\bibitem[\protect\citeauthoryear{Zadeh \bgroup et al\mbox.\egroup
  }{2018b}]{Zadeh2018Multi}
Zadeh, A.; Liang, P.~P.; Poria, S.; Vij, P.; Cambria, E.; and Morency, L.~P.
\newblock 2018b.
\newblock Multi-attention recurrent network for human communication
  comprehension.
\newblock In {\em AAAI},  5642--5649.

\bibitem[\protect\citeauthoryear{Zadeh \bgroup et al\mbox.\egroup
  }{2018c}]{MOSEI}
Zadeh, A.; Liang, P.~P.; Vanbriesen, J.; Poria, S.; Tong, E.; Cambria, E.;
  Chen, M.; and Morency, L.~P.
\newblock 2018c.
\newblock Multimodal language analysis in the wild: Cmu-mosei dataset and
  interpretable dynamic fusion graph.
\newblock In {\em ACL},  2236--2246.

\bibitem[\protect\citeauthoryear{Zeiler}{2012}]{ADADELTA}
Zeiler, M.~D.
\newblock 2012.
\newblock Adadelta: An adaptive learning rate method.
\newblock {\em preprint, arXiv:1212.5701v1}.

\end{thebibliography}

\end{document}